\documentclass[letterpaper, 10 pt, journal, twoside]{IEEEtran}

\IEEEoverridecommandlockouts                              %

\usepackage{amsmath}
\usepackage{mathtools,amssymb}
\usepackage{hyperref}
\usepackage{bm}
\usepackage{color}
\usepackage{booktabs}
\usepackage{multirow}
\usepackage[linesnumbered,ruled,vlined]{algorithm2e}
\usepackage[binary-units]{siunitx}
\usepackage{subcaption}
\usepackage{makecell}
\usepackage{tikz}
\usetikzlibrary{calc}

\newcommand{\algrule}[1][.4pt]{\par\vskip.2\baselineskip\hrule height #1\par\vskip.2\baselineskip}

\newcommand{\cspace}[0]{\mathcal{C}}
\newcommand{\world}[0]{\mathcal{W}}
\newcommand{\Cobs}[0]{\mathcal{C}_{obs}}
\newcommand{\Cfree}[0]{\mathcal{C}_{free}}

\newcommand{\Rthree}[0]{\mathbb{R}^3}
\newcommand{\SOthree}[0]{\mathbb{SO}(3)}
\newcommand{\obstacles}[0]{\mathcal{O}}
\newcommand{\robot}[0]{\mathcal{A}}

\newcommand{\disable}[1]{}
\newcommand{\norm}[1]{\left\lVert#1\right\rVert}

\DeclareMathOperator*{\minimize}{\text{minimize}}
\DeclareMathOperator*{\subjectto}{\text{s.t.}}

\title{Minimum-Time Quadrotor Waypoint Flight\\ in Cluttered Environments}

\author{Robert Penicka and Davide Scaramuzza%
\thanks{Manuscript received: September, 9, 2021; Revised December, 21, 2021; Accepted January, 26, 2022.
This paper was recommended for publication by Editor P. Pounds upon evaluation of the Associate Editor and Reviewers' comments.
This work was supported by the National Centre of Competence in Research (NCCR) Robotics through the Swiss National Science Foundation (SNSF) and the European Union’s Horizon 2020 Research and Innovation Programme under grant agreement No. 871479 (AERIAL-CORE) and the European Research Council (ERC) under grant agreement No. 864042 (AGILEFLIGHT).}
\thanks{The authors are with the Robotics and Perception Group, Department of Informatics, University of Zurich, and Department of Neuroinformatics, University of Zurich and ETH Zurich, Switzerland (\protect\url{http://rpg.ifi.uzh.ch}), {\tt\footnotesize (email: penicka@ifi.uzh.ch)}.}
\thanks{Digital Object Identifier (DOI): see top of this page.}
}

\begin{document}

\maketitle

\begin{abstract}
We tackle the problem of planning a minimum-time trajectory for a quadrotor over a sequence of specified waypoints in the presence of obstacles while exploiting the full quadrotor dynamics.
This problem is crucial for autonomous search and rescue and drone racing scenarios but was, so far, unaddressed by the robotics community \emph{in its entirety} due to the challenges of minimizing time in the presence of the non-convex constraints posed by collision avoidance.
Early works relied on simplified dynamics or polynomial trajectory representations that did not exploit the full actuator potential of a quadrotor and, thus, did not aim at minimizing time.
We address this challenging problem by using a hierarchical, sampling-based method with an incrementally more complex quadrotor model.
Our method first finds paths in different topologies to guide subsequent trajectory search for a kinodynamic point-mass model.
Then, it uses an asymptotically-optimal, kinodynamic sampling-based method based on a full quadrotor model on top of the point-mass solution to find a feasible trajectory with a time-optimal objective.
The proposed method is shown to outperform all related baselines in cluttered environments and is further validated in real-world flights at over 60km/h in one of the world's largest motion capture systems.
We release the code open source.

\end{abstract}


{\small
\vspace{0.1em}
\noindent \textbf{Code:} \url{https://github.com/uzh-rpg/sb_min_time_quadrotor_planning}\\
\noindent \textbf{Video:} \url{https://youtu.be/TIvvHtzRwSo}
\vspace{-0.5em}
}
\section{Introduction}

\IEEEPARstart{Q}{uadrotors} are among the most agile and maneuverable flying machines~\cite{spectrum2020acrobatics}.
This renders them the ideal platform for first responders after disasters, such as earthquakes, forest fires, or floods, to search for survivors as quickly as possible. To push research in this field, autonomous drone racing has emerged as a research field, with international competitions being organized, such  as  the  Autonomous Drone  Racing  series  at  the  recent  IROS  and  NeurIPS  conferences~\cite{moon2019challenges,guerra2019flightgoggles,Madaan20arxiv} and the AlphaPilot challenge~\cite{foehn2020alphapilot,han2021fastracing}.
A key requirement in autonomous drone racing is to plan a trajectory that minimizes the time to fly through a sequence of gates or doorways while avoiding collisions with the environment.
This makes drone racing an ideal benchmark scenario for the humanitarian use of drones in search and rescue.

\begin{figure}[!t]
   \centering
   \includegraphics[width=1.0\columnwidth,trim=0 100 400 0, clip]{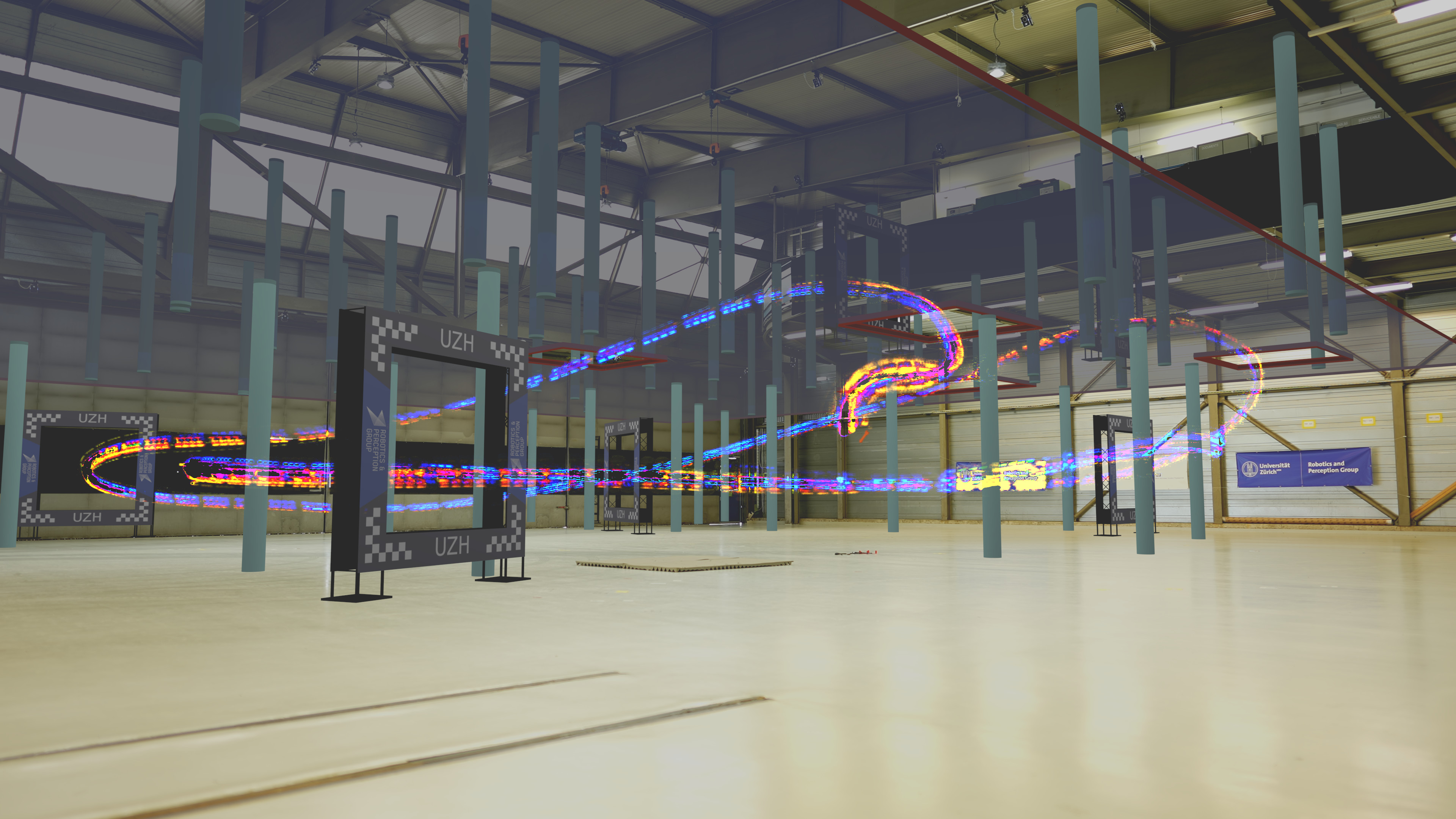}
   \vspace{-1.5em}
   \caption{\label{fig:illustration}
   Our quadrotor executes a near time-optimal multi-waypoint trajectory planned by the proposed method in a racing track with virtual obstacles, reaching speeds over 60km/h.
   The picture was taken in our large indoor drone-testing arena ($30\times30\times8 m^3$) equipped with a motion capture system. The photo was augmented with virtual obstacles simulating a two-floor cluttered environment.
   \vspace{-1.5em}}
\end{figure}

Both drone racing and search and rescue motivate the tackled problem of minimum-time trajectory planning through a sequence of waypoints in cluttered environments.
This problem has not been previously solved in its entirety.
Existing methods either do not consider obstacles in time-optimal planning~\cite{foehn2020cpc}, or cannot use full actuation due to simplification of the quadrotor dynamics~\cite{liu2018search}.
Other methods do not support multi-waypoint scenarios and time-optimal objective~\cite{Webb13_KinodynamicRRTstar}, or rely on polynomials~\cite{richter2016polynomial}, that cannot represent time-optimal trajectories due to their inherent smoothness.

The proposed method uses a hierarchical, sampling-based approach that follows three high-level steps with an incrementally more complex quadrotor model. 
First, topological paths between the individual waypoints are found using a variant of the Probabilistic Roadmap~(PRM)~\cite{TRA96KavrakiPRM}. 
Paths with different homotopy classes (e.g., going from different sides around an obstacle) are found in the roadmap using a combination of iterative shortest path search and roadmap pruning.
In the second step, we use the topological paths to guide trajectory planning for 3D kinodynamic point-mass model.
A trajectory primitive between two waypoints of the point-mass with limited acceleration is found using a gradient descent method.
A dynamic programming graph search algorithm is then employed to find a minimum-time collision-free trajectory for the point-mass by searching over the topological paths in different homotopy classes.
In the final step, a trajectory exploiting the full quadrotor model is planned using a kinodynamic sampling-based Stable Sparse RRT~(SST)~\cite{Li16_sst} algorithm, extended for time-optimal objective and multi-waypoint planning.
The sampling is guided using the point-mass trajectory to effectively explore the quadrotor input and state spaces. %
The proposed hierarchical method differs from existing approaches by considering not only the correct actuation limits and the full quadrotor model but also by planning in presence of obstacles with a minimum-time objective.

To the best of our knowledge, this is the first multi-waypoint quadrotor planner for cluttered environments that focuses solely on the minimum-time objective.
The contributions of this paper are considered as follows.
We propose a new topological path search to find distinct paths through multiple waypoints.
We extend the point-mass trajectory search~\cite{foehn2020alphapilot} to use acceleration norm constraints and account for gravity to better approximate the quadrotor translation dynamics.
Finally, we extend the SST~\cite{Li16_sst} for multi-waypoint scenarios with minimum-time objective, and for sampling guided by the point-mass trajectory.
The proposed hierarchical method is shown to outperform all related baselines in cluttered environments, and is further validated in real-world flights at over 60km/h in one of the world's largest motion capture systems.
The source code is released as an open-source package.

\section{Related Work}\label{sec:related}

The existing approaches for quadrotor trajectory planning can be mostly categorized as polynomial approaches, discrete-time search-based methods, sampling-based approaches, and optimization methods.

The polynomial methods~\cite{richter2016polynomial,burri2015real-time,han2021fastracing,mueller2015TRO_minjerk} represent trajectories as continuous-time polynomials.
They exploit quadrotors' differential flatness property~\cite{IJRR2012mellinger_poly} that allows planning for only four flat outputs, with their high order derivatives, to get the quadrotor states and control inputs. 
The polynomial trajectory planning is widely used for its computational efficiency, however, it can not minimize time by design due to the inherent smoothness of polynomials.
Therefore, polynomial trajectories cannot represent the bang-bang input changes required for using full quadrotor actuation and can reach maximal motor forces only in a limited number of times by either scaling time of the found trajectory~\cite{richter2016polynomial} or by sampling boundary conditions~\cite{mueller2015TRO_minjerk}.

Other methods~\cite{Zhou2019ral_traj_gen_bspline,Zhou2020_guided_gradient_planning,zhou2020raptor,Penin2018RAL_vision_reactive_planning} that also leverage the differential flatness property use B-spline representation for trajectory optimization.
These methods jointly optimize smoothness, dynamic feasibility, collision cost, and also safety~\cite{zhou2020raptor} or visual tracking objectives~\cite{Penin2018RAL_vision_reactive_planning}.
The computational efficiency of the methods is suitable for online replanning, however, they can not be used for minimum-time flight due to the different objectives.
Furthermore, using full actuation of the quadrotor is limited as the dynamic feasibility is enforced as a soft constraint and mostly in a per-axis form.
Similarly to our method, \cite{zhou2020raptor,Zhou2019ral_traj_gen_bspline} use topological path search to find distinct paths, however, using a different algorithm with guard nodes in PRM.

The search-based methods~\cite{Liu_search_based_LQMTC,liu18_searchbased} use discrete-time and discrete-state representations and convert the trajectory planning to a graph search problem.
The search-based approaches optimize time up to discretization, however, they suffer from the curse of dimensionality.
Furthermore, the employed per-axis acceleration limits have to be set pessimistically to plan feasible but suboptimal trajectories for the full quadrotor model. 
The existing search-based methods support planning only between two states, and extension to multi-waypoint scenarios would be challenging with additional search dimensions for already computationally demanding methods.

The sampling-based kinodynamic methods~\cite{Li16_sst,Hauser2016TRO_state_cost_space} plan for a dynamic system like quadrotor by growing a tree of states where in each iteration a random state is selected and forward integrated using random inputs for a random time.
Alternatively, RRT* algorithm~\cite{Webb13_KinodynamicRRTstar} can be used for linear systems or linearized quadrotor model around hover conditions, however, this prohibits fast and aggressive flights.
Other methods randomly sample polynomial primitives~\cite{Zhiling_RRTstar_min_jerk} or states of simplified quadrotor kinodynamic point-mass model~\cite{foehn2020alphapilot}.
All above sampling-based methods, with the exception of~\cite{foehn2020alphapilot}, focus on minimizing trajectory length. 
This requires only a slight deviation from the quadrotor hover inputs.
However, the time-optimal inputs lie on the boundary of the input space, which is more challenging to sample using uniform random sampling.
Furthermore, the typically much faster rotational dynamics of the quadrotor, compared to the translational dynamics, render the existing sampling-based planners with Voronoi bias~\cite{Li16_sst,Hauser2016TRO_state_cost_space} unusable for the minimum-time objective.
The Voronoi bias causes preferred expansion of states on the boundary of already explored space, which can block the planning algorithms when a state on the boundary is repeatedly selected and expanded to the unfeasible state, e.g., due to the quadrotor's rotation speed constraints.
This motivated the proposed method to use the hierarchical approach to guide quadrotor input sampling using the previously found point-mass solution.
Such point-mass solution has approximate information about the state and input of the full quadrotor dynamics, and can be used to effectively decrease the sampling complexity and mitigate the effects of the Voronoi bias.

Finally, a method that solves the most similar problem is the optimization-based planning for quadrotor waypoint flight~\cite{foehn2020cpc}.
The method finds time-optimal trajectories for the full quadrotor model through a sequence of waypoints, however, it does not consider environments cluttered with obstacles.
This renders the method mostly unusable when additional non-convex constraints are introduced to prevent collisions.
The proposed sampling-based method is able to plan minimum-time trajectories for the cluttered environments by checking collisions in all levels of hierarchical planning and by guiding the planning using already found collision-free paths from the previous level.

\section{Problem Statement\label{sec:problem}}

We formulate the tackled minimum-time motion planning problem through a sequence of waypoints in a cluttered environment using the classical notion of the configuration space $\cspace$~\cite{lavalle2006planning}.
Having the world $\world = \Rthree$ with the obstacles $\obstacles = \{\obstacles_{1},\dots,\obstacles_{m}\} \subset \world$, the motion planning problem is to determine a collision free-path for a robot $\robot \subset \world$ between two locations in $\world$ such that the path avoids~$\obstacles$.
In this paper, multiple waypoints $P_w = \{\bm{p}_{wi}, i \in [1,\ldots,N]\}$ have to be visited with certain proximity $r_{tol}$ in a given order, e.g. corresponding to the gate centers in the drone racing.

Let $\robot(\bm{x}) \subset \world $ denote the geometry of the robot at a configuration $\bm{x}$.
The robot  can move in free space $\Cfree = \cspace \setminus \Cobs$, where $\Cobs = \{ \bm{x} \in \cspace | \delta(\robot(\bm{x}),\obstacles) \leq d_c \} \subseteq \cspace$ is a set of configurations where the robot is in collision, i.e. the shortest distance $\delta(\cdot,\cdot)$ between the robot and any obstacle is below a given threshold $d_c$.
For the considered quadrotor, the robot's configuration $\bm{x}=\begin{bmatrix} \bm{p},\bm{q},\bm{v},\bm{\omega} \end{bmatrix}^{T}$ consists of its position $\bm{p} \in \Rthree$, velocity $\bm{v} \in \Rthree$, unit quaternion rotation $\bm{q} \in \SOthree$, and body rates $\bm{\omega} \in \Rthree$. 
The dynamic equations with total collective thrust $\bm{f}_{T}$ and body torque $\bm{\tau}$ inputs are
\begin{align}
\label{eq:quat_dyn}
  \begin{aligned}
    \bm{\dot{p}} &= \bm{v} \vphantom{\frac{1}{2}} \\
    \bm{\dot{v}} &= \frac{1}{m}R(\bm{q})\bm{f}_{T} + \bm{g}
  \end{aligned}
  &&
  \begin{aligned}
    \bm{\dot{q}} &= \frac{1}{2} \bm{q} \odot \begin{bmatrix} 0 \\ \bm{\omega} \end{bmatrix} \\
    \bm{\dot{\omega}} &= \bm{J}^{-1} (\bm{\tau} - \bm{\omega} \times \bm{J} \bm{\omega}) \vphantom{\frac{1}{2}} \text{,}
  \end{aligned}
\end{align}
where $\odot$ denotes quaternion multiplication, $R(\bm{q})$ quaternion rotation, $m$ quadrotor mass, $\bm{J}$ its inertia, and $\bm{g}$ is gravity.

However, the real quadrotor inputs are single rotor thrusts $\begin{bmatrix}f_1,f_2,f_3,f_4\end{bmatrix}$ which are used to calculate $\bm{f}_{T}$ and $\bm{\tau}$ as
\begin{equation}
\label{eq:tau_thrust}
\bm{f}_{T}=\begin{bmatrix} 0 \\ 0 \\ \sum f_i \end{bmatrix} \text{, }
\bm{\tau}_{b} = 
\begin{bmatrix} 
 l/\sqrt{2}(f_{1}-f_{2}-f_{3}+f_{4})   \\
 l/\sqrt{2}(-f_{1}-f_{2}+f_{3}+f_{4})   \\
 \kappa (f_{1}-f_{2}+f_{3}-f_{4})
\end{bmatrix} 
\text{,}
\end{equation}
using torque constant $\kappa$ and arm length $l$.
The single rotor thrusts are further constrained~\eqref{eq:motor_constraints} by minimal $f_{min}$ and maximal $f_{max}$ values.
The body rates are limited~\eqref{eq:rate_constraints} by a per-axis maximal allowed value $\omega_{max}$.
\begin{align}
  f_{min} \leq &f_{i} \leq f_{max} \text{, for } i \in \{1,\ldots,4\} \label{eq:motor_constraints}\\
  -\omega_{max} \leq &\omega_{i} \leq \omega_{max} \text{, for } \bm{\omega}={\begin{bmatrix}\omega_{1},\omega_{2},\omega_{3}\end{bmatrix}}^{T} \label{eq:rate_constraints}
\end{align}

The solution to the studied multi-waypoint motion planing between specified start $\bm{x}_s \in \Cfree$ and end $\bm{x}_e \in \Cfree$ configurations is a continuous sequence of trajectories $\tau_i: [0,1] \to \Cfree$ for $i \in \{0,\ldots,N\}$ that pass through $P_w$. 
The sequence has the given start $\tau_{0}(0) = \bm{x}_{s}$ and end $\tau_{N}(1) = \bm{x}_{e}$, the initial positions of the trajectories $\tau_{i}(0)_{p}$ have to be in waypoints' proximity $\norm{\tau_{i}(0)_{p} - \bm{p}_{wi}} \leq r_{tol}$ for $i \in \{1,\ldots,N\}$, and the sequence has to be continuous $\tau_{i-1}(1) = \tau_{i}(0)$ for $i \in \{1,\ldots,N\}$.
The objective of the problem is then to minimize the time $T=\sum_{i=0}^{N} t_i$ of reaching the end configuration, where $t_i$ corresponds to the time duration of $\tau_{i}$ given \eqref{eq:quat_dyn}-\eqref{eq:rate_constraints}.
The whole multi-waypoint minimum-time planning problem can be summarized as
\begin{equation} \label{objective}
   \begin{split}
      \minimize_{\tau_0 \ldots \tau_N}~& T=\sum_{i=0}^{N} t_i    \\
      \subjectto \text{ }& \tau_i \in \Cfree \text{ for } i \in \{0,\ldots,N\}\text{,}\\
      & \tau_{0}(0) = \bm{x}_{s} \text{, } \tau_{N}(1) = \bm{x}_{e}\text{,}\\
      & \norm{\tau_{i}(0)_{p} - \bm{p}_{wi}} \leq r_{tol}\text{ for } i \in \{1,\ldots,N\}\text{,}\\
      & \tau_{i-1}(1) = \tau_{i}(0) \text{ for } i \in \{1,\ldots,N\} \text{,}\\
      & \eqref{eq:quat_dyn}\text{, } \eqref{eq:tau_thrust}\text{, } \eqref{eq:motor_constraints}\text{, } \eqref{eq:rate_constraints} \text{.}
   \end{split}
\end{equation}

\section{Method\label{sec:method}}

\begin{figure*}[!ht]
   \centering
   \setlength{\tabcolsep}{0.2em}
  \begin{tabular}{ccc}
  \subcaptionbox{Distinct paths\label{fig:stages_roadmap}}{\includegraphics[height=0.16\linewidth,width=0.3\linewidth]{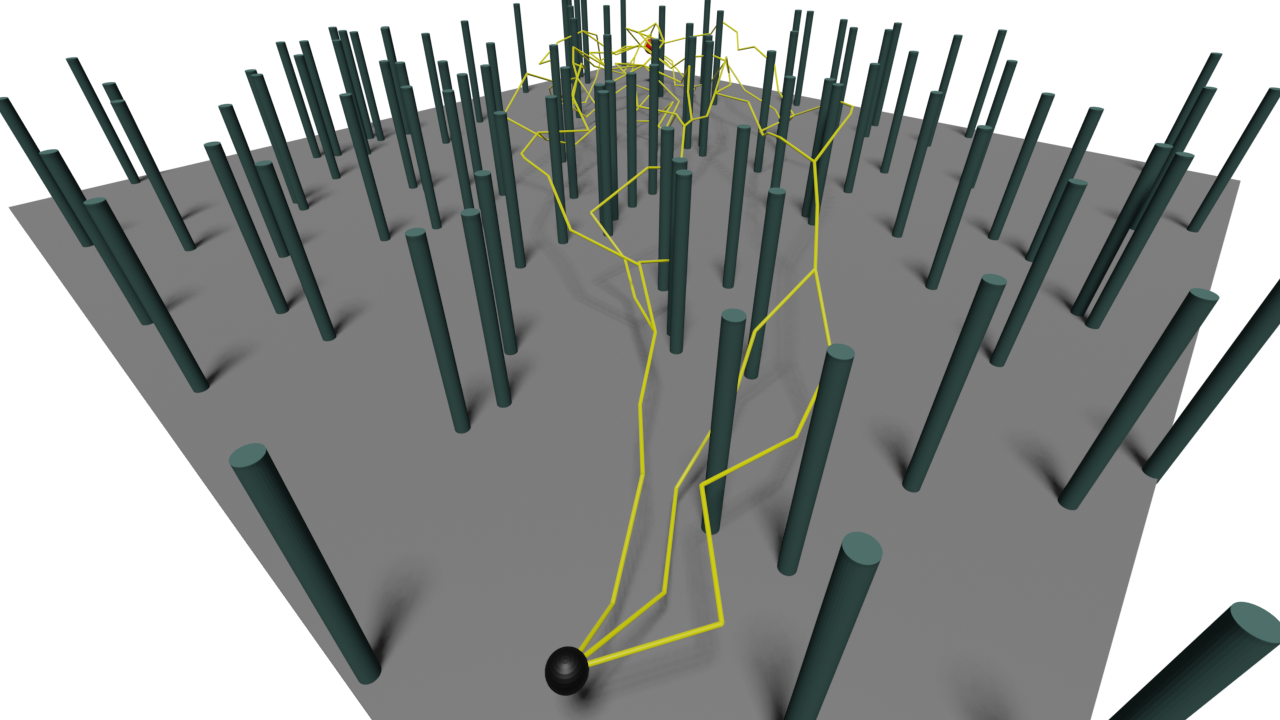}\vspace{-0.6em}}   &
  \subcaptionbox{Shortened paths\label{fig:stages_shortened}}{\includegraphics[height=0.16\linewidth,width=0.3\linewidth]{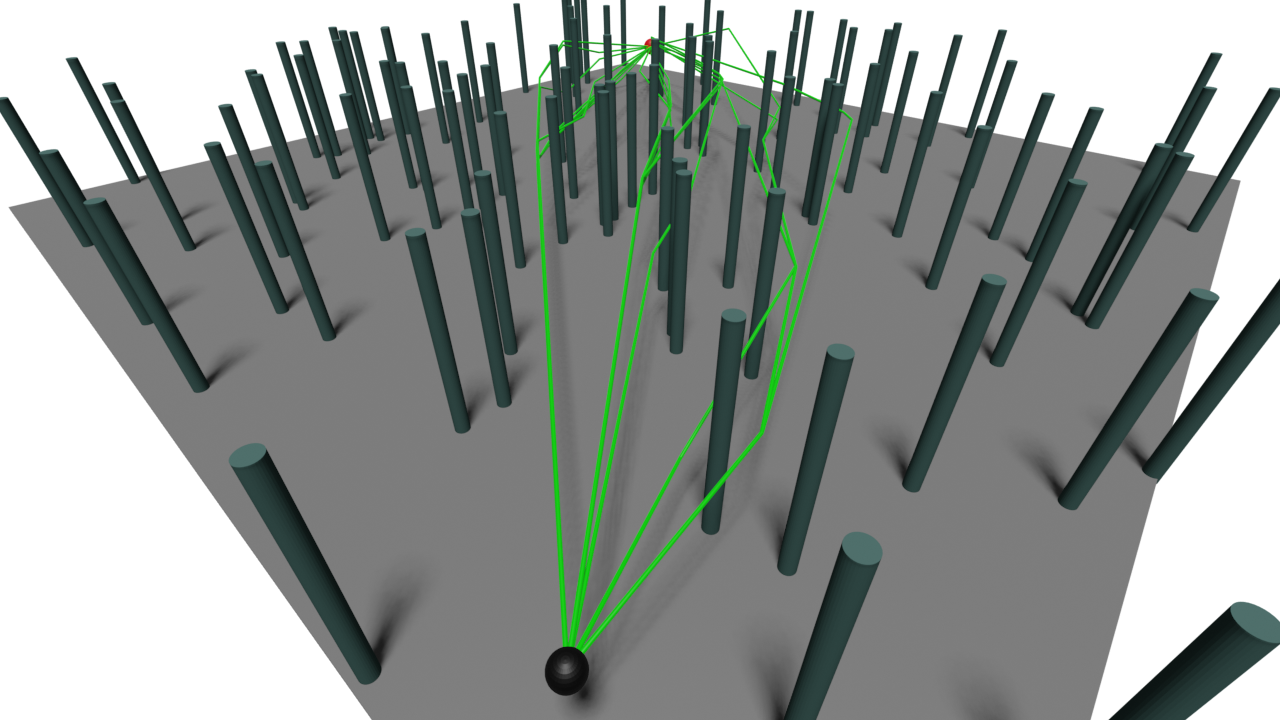}\vspace{-0.6em}}   &
  \subcaptionbox{Filtered unique paths\label{fig:stages_filtered}}{\includegraphics[height=0.16\linewidth,width=0.3\linewidth]{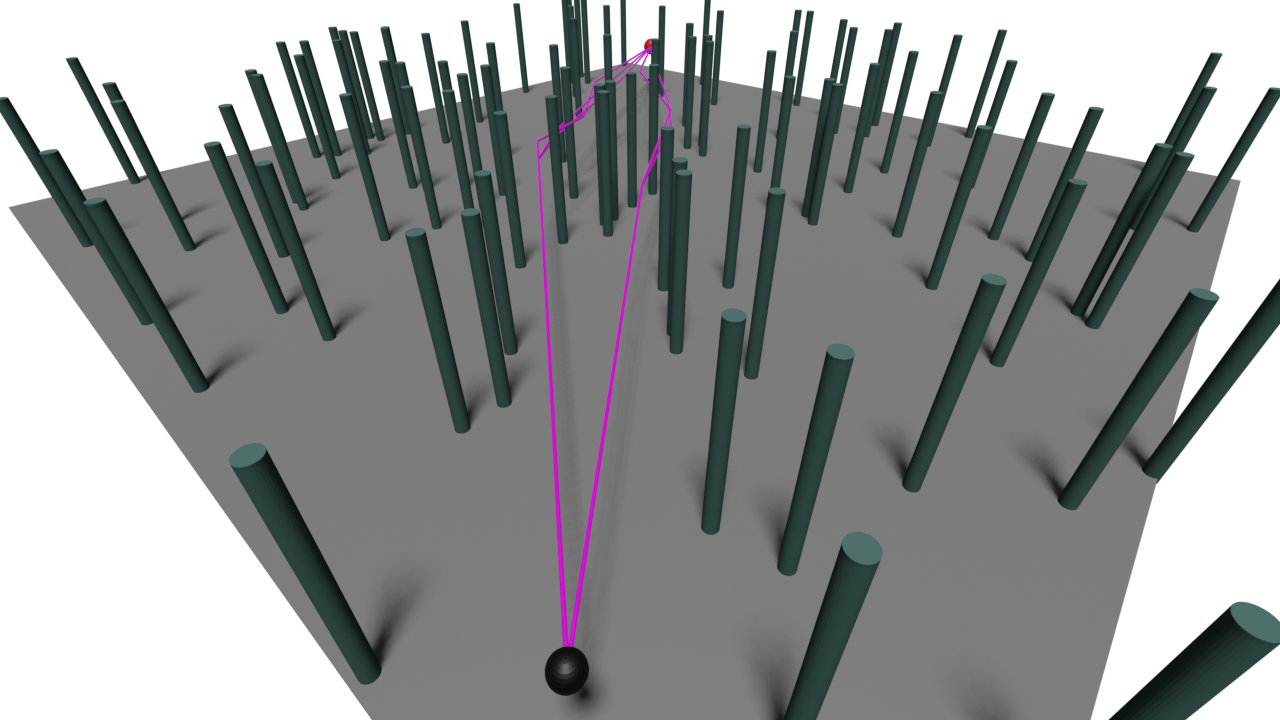}\vspace{-0.6em}}  \\
  \subcaptionbox{Point-mass solution\label{fig:stages_pmm}}{\includegraphics[height=0.16\linewidth,width=0.3\linewidth]{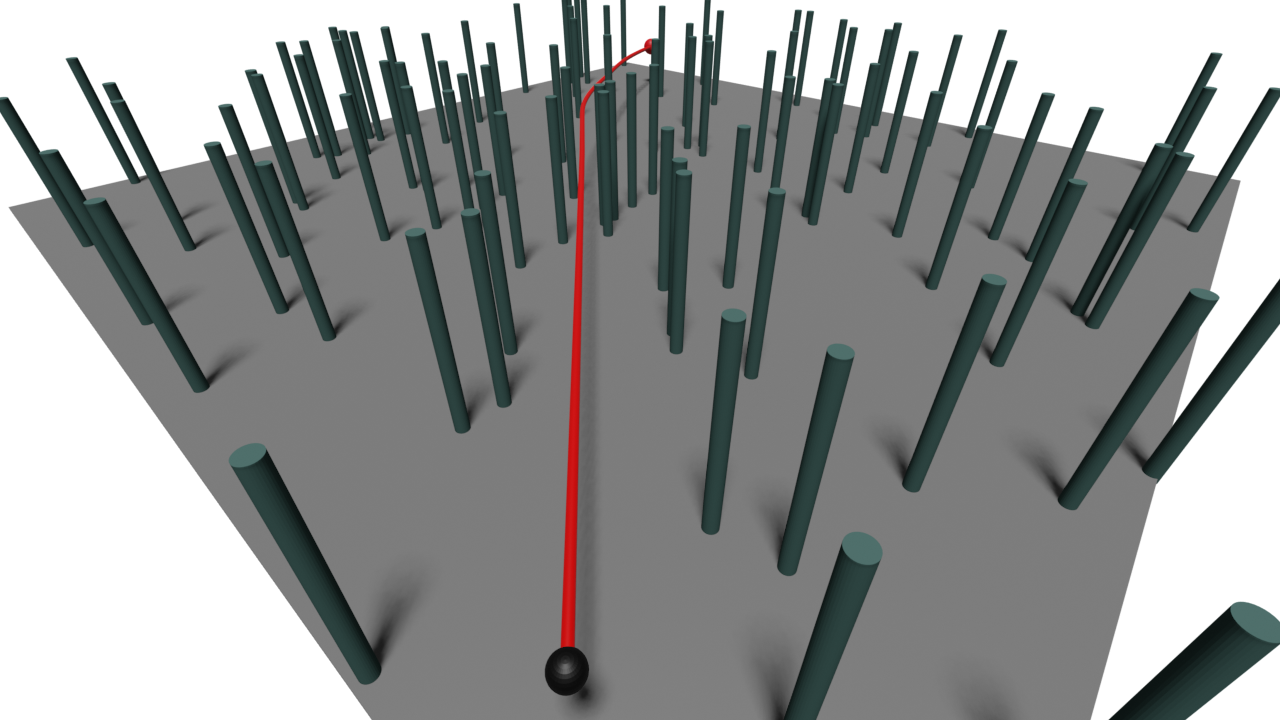}\vspace{-0.6em}}   &
  \subcaptionbox{Planning tree\label{fig:stages_tree}}{\includegraphics[height=0.16\linewidth,width=0.3\linewidth]{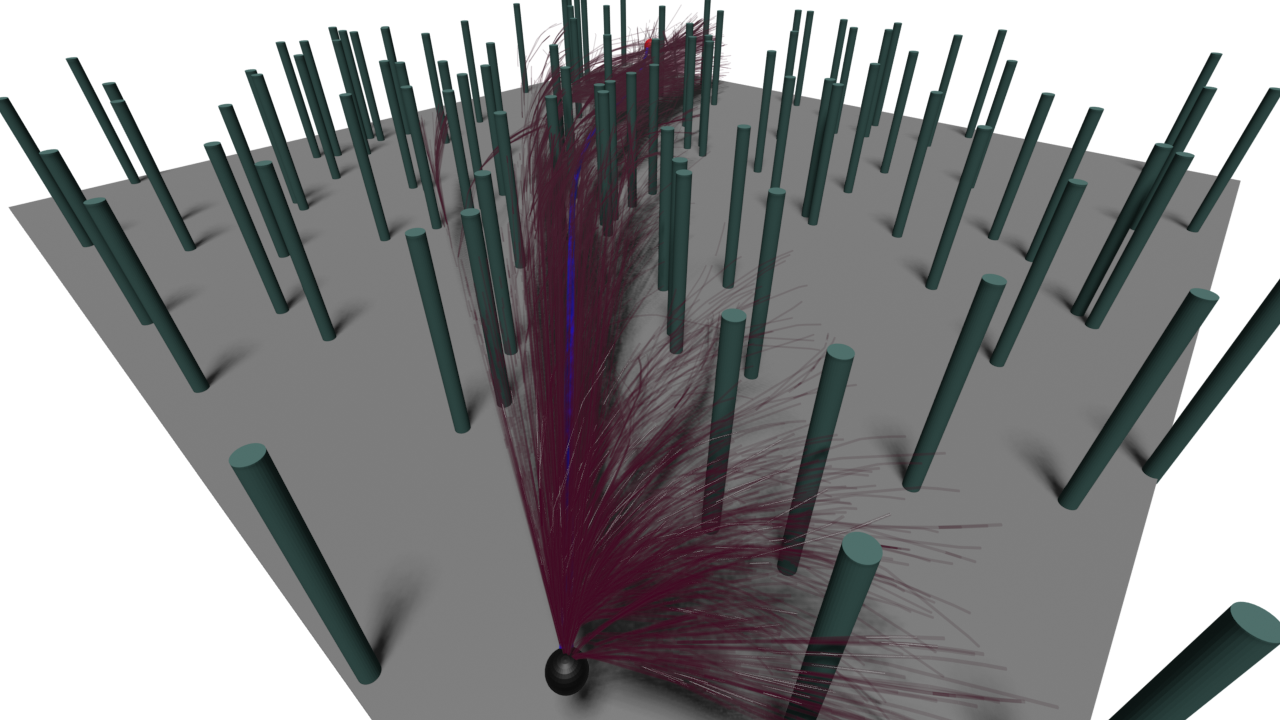}\vspace{-0.6em}}   &
  \subcaptionbox{Final trajectory\label{fig:stages_final}}{\includegraphics[height=0.16\linewidth,width=0.3\linewidth]{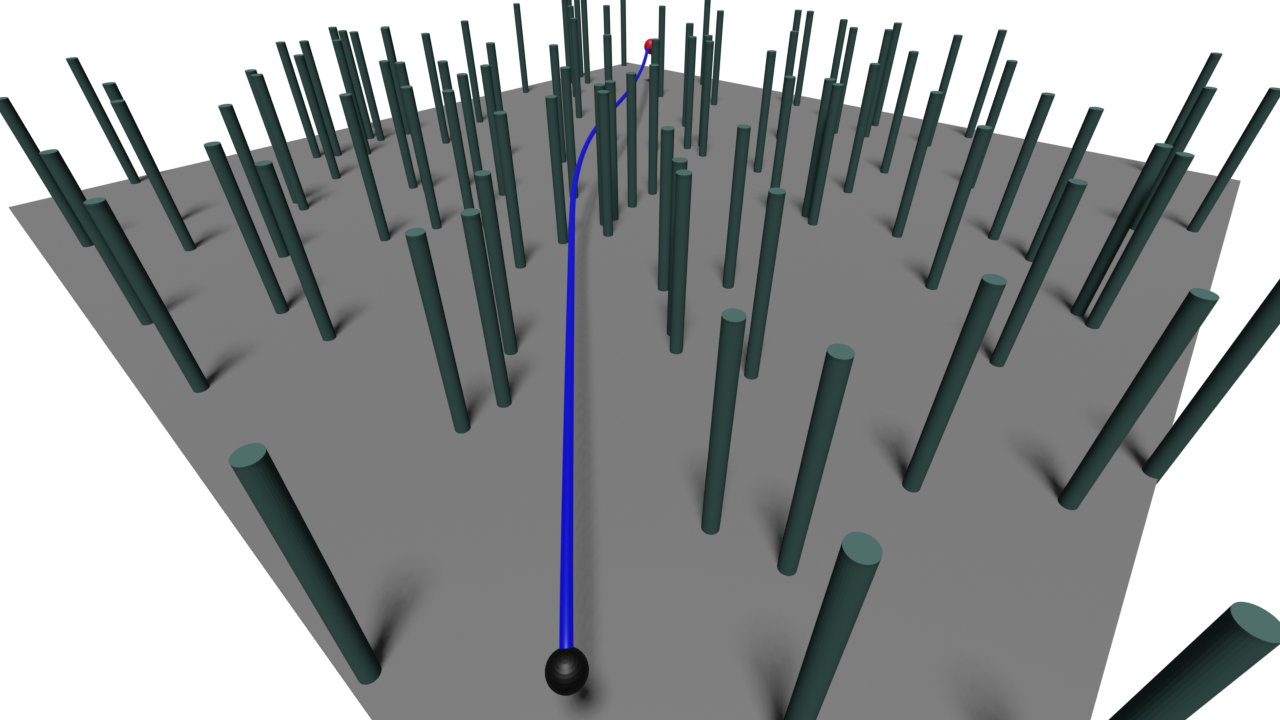}\vspace{-0.6em}} 
   \end{tabular}
   \vspace{-0.6em}
   \caption{
   Visualization of the individual stages of the hierarchical approach.
   Distinct topological paths are firstly found between goals (a) and are consequently shortened (b).
   Filtered paths (c) are then used to find the minimum-time trajectory for the point-mass model (d).
   Finally, the sampling-based method uses the point-mass solution to guide exploration of the solution space (e) to find the final trajectory (f).
   The start and goal are colored black and red, respectively.
      \label{fig:method} 
   \vspace{-1.7em}
    }
   
\end{figure*}

The proposed method uses a hierarchical approach that initially finds diverse topological paths with a variant of Probabilistic Roadmap~(PRM)~\cite{TRA96KavrakiPRM}.
Then, the fastest trajectory for 3D point-mass model is found by searching over the topological paths.
Finally, the SST method is used to find a minimum-time trajectory over multiple waypoints for the full model of the quadrotor using the point-mass solution to guide the exploration.
The whole method uses Euclidean Signed Distance Field~(ESDF) as a priory known map of the environment.
The method is visualized in Figure~\ref{fig:method}.

\subsection{Topological path search\label{subsec:topo_paths}}

The topological path search is used to find distinct paths with different homotopy classes, e.g. passing around an obstacle from a different side.
This is necessary as the typically sought minimum-length path does not necessarily translate to the minimum-time trajectory for the dynamic point-mass model and further for the full quadrotor model.

The method described in Algorithm~\ref{alg:topo_PRM} plans sequentially for pair of adjacent goals in $G=[\bm{p}_s, \bm{w}_{wi} \in P_w , \bm{p}_e]$, where $\bm{p}_s$ and $\bm{p}_e$ are positions of the start and goal configurations, respectively.
The PRM is used to create a roadmap $(V,E)$ by sampling ellipsoid between two goals similarly to the Informed-RRT$^*$ \cite{Gammell14_informedrrtstar}. 
The number of samples and ellipsoid major axis are iteratively increased by a constant factor if no path is found.
Once at least one path can be found in the roadmap, method \textbf{FindDistinctPaths} is used to find distinct paths between the goals as described in Alg.~\ref{alg:topo_distinct_paths}.

The distinct paths are further shortened (similarly to~\cite{zhou2020raptor}) in forward and backward pass.
Afterward, paths with the same homotopy class are removed.
The homotopy is approximated using \emph{uniform visibility deformation}~(UVD)~\cite{zhou2020raptor}, where trajectories $\tau_{i}^{1}:[0,1] $, $\tau_{i}^{2}:[0,1]$ belongs to same UVD class if the straight line connection $\tau_{i}^{1}(s)\rightarrow\tau_{i}^{2}(s)$ is collision free for all $s \in [0,1]$.
Finally, the paths are filtered by keeping only paths shorter than a fixed factor of the shortest one and by limiting the number of totally kept paths.
Additionally, paths with the wrong angle of reaching a waypoint, i.e., the direction of flying through a gate, can be filtered out. 

\SetAlFnt{\footnotesize}
\setlength{\textfloatsep}{0pt}
\begin{algorithm}[!htb]
   \caption{Topological PRM\label{alg:topo_PRM}} 
   \SetKwInOut{Input}{Input}
   \SetKwInOut{Output}{Out}
   \Input{$G$ -- goals} 
   \Output{$\Pi_0, \ldots, \Pi_{N}$ -- found topological paths}
   \algrule
   \DontPrintSemicolon
   \ForEach{$ i \in 1,\ldots,\norm{G}-1 $}{
    $(V,E) \leftarrow \textbf{PRM}(G[i-1],G[i])$\\
    $\Pi_{i-1}^{d} \leftarrow \textbf{findDistinctPaths}((V,P),\{G[i-1]\},G[1])$\\
    $\Pi_{i-1}^{*} \leftarrow \textbf{shortenPaths}(\Pi_{i-1}^{d})$\\
    $\Pi_{i-1}^{u} \leftarrow \textbf{removeEquivalentPaths}(\Pi_{i-1}^{*})$\\
    $\Pi_{i-1} \leftarrow \textbf{filterPaths}(\Pi_{i-1}^{u})$\\
   }
\vspace{-0.5em}
\end{algorithm}

The distinct path search in Algorithm~\ref{alg:topo_distinct_paths} iteratively searches for the shortest path using Dijkstra's algorithm between given start nodes $S$ and end nodes $F$ in the roadmap.
For each path, a node $v_n$ with smallest clearance $d_{n}$ from obstacles is found.
Additionally, nodes $V_n$ that are connectable to $v_n$ using collision-free path with length within $d_{n}$ are found.
Nodes $V_n$ and $v_n$ are then removed from the roadmap and saved among deleted nodes $V_d$.
The shortest path search is repeated until no new path can be found.
This is either if no alternative path exists or vertices in some narrow passages were removed.
To address the second case, the algorithm is recursively called from both $S$ to $V_d$ and from $V_d$ to $F$.
Finally, paths between original $S$ and $F$ nodes are connected using the found paths with corresponding nodes in $V_d$.

\SetAlFnt{\footnotesize}
\setlength{\textfloatsep}{0pt}
\SetKwRepeat{Do}{do}{while}
\vspace{-0.75em}
\begin{algorithm}[!htb]
    \caption{Distinct Path Search\label{alg:topo_distinct_paths}} 
    \SetKwInOut{Input}{Input}
    \SetKwInOut{Output}{Out}
    \Input{$(V,E)$ -- roadmap, $S$ -- start nodes, $F$ -- end nodes} 
    \Output{$\Pi$ -- found distinct topological paths}
    \algrule
    \DontPrintSemicolon
    $V_d = \emptyset$ \tcp*{set of deleted nodes} 
    \Do{$\Pi_s \neq \emptyset$}{
          $\Pi \leftarrow \Pi \cup \leftarrow \textbf{findShortestPaths}((V,E),S,E)$\\
          \ForEach{$\pi_s \in \Pi_s$}{
              $v_n, d_n \leftarrow \textbf{nearestNodeToObstacle}(\pi_s)$\\
              $V_n \leftarrow \{v \in V \mid \norm{v_n-v} \leq d_n-d_c \} $\label{alg:tdp_line_near nodes}\\
              $V = V \setminus (\{v_n\} \cup V_n)$ \label{alg:tdp_line_new_nodes}\\
              $V_d \leftarrow V_d \cup \{v_n\} \cup V_n$\label{alg:tdp_line_deleted}\\
          }
    }
    \If{$V_d \neq \emptyset$}{
       $\Pi_b \leftarrow \textbf{FindDistinctPaths}((V,P),S,V_d)$\\
       $\Pi_a \leftarrow \textbf{FindDistinctPaths}((V,P),V_d,E)$\\
       $\Pi \leftarrow \Pi \cup \textbf{connectPaths}(\Pi_b,\Pi_a,V_d)$
    } 
\vspace{-0.5em}   
\end{algorithm}
\vspace{-1.0em}

\subsection{Point-mass trajectory planning}

The found topological paths are further used to plan a trajectory for 3D point-mass model that approximates the translation part of the quadrotor dynamics.
A minimum-time motion primitive is created between any two states of the point-mass model.
A dynamic programming search, that utilizes the primitive, is then used to find a collision-free point-mass trajectory between multiple waypoints.
The distinct topological paths are used during the search to constraint the position of the primitives to the already found collision-free topological paths.

\subsubsection{Point-mass motion primitive}

We use the 3D point-mass model with state consisting of position $\bm{p}$ and velocity $\bm{v}$, and with input on acceleration $\bm{u} = \bm{a}$.
The motion primitive is a trajectory from given start position $\bm{p}^s = (p^s_1,p^s_2,p^s_3)^T$ and start velocity $\bm{v}^s = (v^s_1,v^s_2,v^s_3)^T$ to end position $\bm{p}^e = (p^e_1,p^e_2,p^e_3)^T$ and end velocity $\bm{v}^e = (v^e_1,v^e_2,v^e_3)^T$.
For a single axis, without loss of generality for the $x$-axis, it could be shown that for dynamics $\ddot{p}_1(t)=u_1(t)$ the time-optimal control input using Pontryagin's Maximum Principle (PMP) is a bang-bang using a given maximal per-axis acceleration $a_{1}$.
The optimal control has form
\begin{equation}
\label{eq:bang-bang}
a_{1}^* = \begin{cases}
      a_{1}  & 0 \leq t \leq t_1^* ,\\   
      -a_{1} & t_1^* \leq t \leq T_1^* ,
   \end{cases}
\end{equation}
or with switched inputs of $-a_{1}$ followed by $a_{1}$.
The minimum final time $T_1^* = T_{1}^{*}(a_{1},p^s_1,v^s_1,p^e_1,v^e_1)$ can be found in closed-form as a solution to a set of four classical kinematic equations that describe position and velocity for both bang-bang parts.
The only unknowns are then the $t_1^*$, $T_1^*$, and the position and velocity in $t_1^*$.

A similar approach was proposed in~\cite{foehn2020alphapilot} where the authors used per-axis acceleration limits. 
In this work, the point-mass primitive is extended by constraining the acceleration norm $a_{max}=4 f_{max} / m$ over joint axes and by accounting for gravity. 
This corresponds to point-mass thrust acceleration $\bm{a}_{t} = (a_{t,1},a_{t,2},a_{t,3})^{T}, \norm{\bm{a}_{t}}=a_{max}$ with body acceleration used for the per axis closed-form solutions in \eqref{eq:bang-bang} as $\bm{a} = \bm{a}_{t} + \bm{g}$.
Finally, to minimize the time $T^{*}(\bm{a},\bm{p}^s,\bm{v}^s,\bm{p}^e,\bm{v}^e)=\max_{i=\{1,2,3\}} T_{i}^{*}(a_{i},\ldots)$ of the coupled 3D point-mass primitive, we optimize over the thrust acceleration $\bm{a}_{t}$ as
\begin{equation} 
\begin{split}
\minimize_{\bm{a_t}}~& T^{*}(\bm{a},\bm{p}^s,\bm{v}^s,\bm{p}^e,\bm{v}^e)\\
\subjectto \text{ }& \bm{a} = \bm{a}_{t} + \bm{g}\text{,} \norm{\bm{a}_{t}}=a_{max}\text{.}\\
\end{split}
\end{equation} 

We use a Gradient Descent (GD) on a sphere with radius $a_{max}$ and project the gradient onto the sphere to keep maximal acceleration norm.
The gradient of individual axes can be found in closed-form similarly to the solution of~\eqref{eq:bang-bang}.
After the GD, the primitive is calculated by stretching the times of the axes $T_{i}^{*}$ to match time $T^{*}$, which can be done in closed-form by lowering the per-axis accelerations, and is necessary to keep the desired $\bm{p}^e$ and $\bm{v}^e$.

\subsubsection{Multi-waypoint point-mass trajectory search}
The point-mass motion primitive is further used to plan a multi-waypoint trajectory utilizing the previously found topological paths $\Pi$.
As the positions are known, the search converts to finding velocities for a given sequence of positions, initially $G=(\bm{p}_s,\bm{p}_{w1},\ldots,\bm{p}_{wN},\bm{p}_e)$, that minimize the time of reaching $\bm{p}_e$.
In~\cite{foehn2020alphapilot} this task is solved (without considering obstacles) by randomly sampling velocities in $G$ with consequent shortest-time search in the computed graph of motion primitives.
This, however, has quadratic complexity in the number of velocity samples per position.
Therefore, we opted for deterministic sampling with a small number of velocity samples, i.e., having only 27 velocity samples in a cone with $3\times3\times3$ samples corresponding to yaw, pitch, and velocity norm dimensions.
Dijkstra’s algorithm is used to find the shortest-time path in such a graph using the point-mass motion primitive between individual samples.
Afterward, the cones are refocused around the found shortest-time path, and the search is repeated until the time converges.
When refocused, the cones are adjusted individually for each dimension.
The samples are either re-centered if a boundary sample is used, or the distance between samples is halved if the center sample is used.
This way, the number of primitive evaluations is limited while iteratively focusing the search toward the minimum-time solution.

The above described \textbf{velocitySearch} method computes point-mass trajectory over the positions in $G$ without considering the obstacles.
An additional graph search over the topological paths, described in Algorithm~\ref{alg:point_mass_tr_search}, is used to find a collision-free point-mass trajectory.
It uses dynamic programming search with priority queue to always test collisions in the currently shortest-time trajectory.
In case of a collision, a new position is added to $G$ between nodes where the first collision occurred. 
A new sequence with an additional position is created for each possible topological path between such nodes.
This way, the algorithm keeps attaching the point-mass trajectory to  the  collision-free topological paths until the shortest feasible path is found.

\SetAlFnt{\footnotesize}
\setlength{\textfloatsep}{0pt}
\begin{algorithm}[!htb]
    \caption{Point-mass trajectory search\label{alg:point_mass_tr_search}} 
    \SetKwInOut{Input}{Input}
    \SetKwInOut{Output}{Out}
    \Input{$\Pi$ -- topological paths, $G$ -- goals} 
    \Output{$\tau_{pm}$ -- found point-mass trajectory}
    \algrule
    \DontPrintSemicolon
    $\pi_{i} \leftarrow \textbf{velocitySearch}(G)$ \tcp*{initial path} 
    $\Pi_h \leftarrow {(\pi_{i},G)}$ \tcp*{initialize heap} 
    \While{$\Pi \neq \emptyset$}{
          $(\pi_b,G_b) \leftarrow \textbf{findAndRemoveBestTrajectory}(\Pi_h)$ \\ %
          \If{$\pi_b$ is collision-free}{
            $\tau_{pm} \leftarrow \pi_b$; break \tcp*{trajectory found} 
          }
          \Else{
          $g_b,g_a \leftarrow \textbf{getNodesOfFirstCollision}(\pi_b)$ \\
          \For{$\pi$ between $g_b,g_a$ found in $\Pi$}{
            $G_{new} \leftarrow \text{ add goal between } g_b,g_a \text{ in } \pi \text{ to }G_b$\\
            $\pi_{new} \leftarrow \textbf{velocitySearch}(G_{new})$\\
            $\Pi_h \leftarrow \Pi_h \cup {(\pi_{new},G_{new})}$\\
          }
          }
    } 
\vspace{-0.5em}
\end{algorithm}
\vspace{-1.0em}

\subsection{Full quadrotor model sampling-based planning}

This part extends the original kinodynamic sampling-based method called SST~\cite{Li16_sst} to the multi-waypoint scenario guided by the point-mass trajectory.
The minimum-time cost is employed instead of the easier-to-sample shortest-path objective used in the original method.
The SST is implemented as the final planning step that uses the full quadrotor dynamics and creates a feasible solution compared to the point-mass approximation.
The guiding with the point-mass solution is, however, used not only to reduce the volume of sampled configuration space but is needed to avoid blocking the SST by Voronoi bias when sampling on the edge of feasible input space for the minimum-time objective.

The SST works by protecting the configuration space with a set of witnesses $S_i$, each containing its center configuration and a representative configuration that contains the node with the best cost within $\delta_{s}$ radius of the center.
A new node is added to the SST tree and to active nodes $V_{a,i}$ only if it reaches unexplored space and forms a new witness, or its cost is better than the cost of its witness representative.
The old representative is replaced in the latter case by the better one and moved to inactive nodes $V_{u,i}$.
This way, the SST keeps only minimum-time trajectories and explores the configuration space until the desired goal is reached.
We extend the original SST to multi-waypoint scenarios by considering $S_i$, $V_{a,i}$ and $V_{u,i}$ for all $i=0,\ldots,\norm{G}-2$, i.e. goals in the sequence $G$.
The planning tree starts in $x_{s}$, and in each iteration, a random goal is selected among the already reached goals.
The tree is allowed to grow to the next goal if the previous goal is reached with tolerance $r_{tol}$.
The multi-waypoint trajectory is thus planned using one tree that keeps expanding towards unreached goals until the stopping conditions are met.

The proposed guided multi-goal SST (Algorithm~\ref{alg:multigoal_sst}) uses the point-mass trajectory $\tau_{pm}$ to create a non-continuous trajectory primitive that has the translational dynamics part of $\tau_{pm}$.
The rotational parts are added to match the acceleration changes of the point-mass using PMP bang-bang or bang-singular-bang rotational inputs.
The inputs are found by calculating quaternion errors in the body x-y plane in the changes of the point-mass accelerations.
The error is used to find rotational axis and maximal angular acceleration producible given the motor constraints~\eqref{eq:motor_constraints}.
Finally, the rotational parts along the found axis are calculated similarly to single-axis motion~\eqref{eq:bang-bang} using the maximal angular acceleration, zero initial and final angular velocities, and angular position from the error quaternions.
The bang-singular-bang solution is used to satisfy the angular velocity constraints~\eqref{eq:rate_constraints} and is calculated similarly to~\eqref{eq:bang-bang} in closed-form.
This trajectory primitive assumes decoupled rotational and translational dynamics, however, it serves as a guiding reference for the SST that uses the full quadrotor dynamics and couples the dynamics while randomizing quadrotor inputs around the primitive.

\SetAlFnt{\footnotesize}
\setlength{\textfloatsep}{0pt}
\begin{algorithm}[!htb]
    \caption{Guided multi-goal SST\label{alg:multigoal_sst}} 
    \SetKwInOut{Input}{Input}
    \SetKwInOut{Output}{Out}
    \Input{$\tau_{pm}$ -- point-mass trajectory, $G$ -- goals} 
    \Output{$\tau$ -- found quadrotor trajectory}
    \algrule
    \DontPrintSemicolon
    $V_{a,i} \leftarrow \emptyset \text{, } V_{u,i} \leftarrow \emptyset \text{, } S_i \leftarrow \emptyset \text{, } \forall i=0,\dots,\norm{G}-2$ \\
    $V_{a,0} \leftarrow \{x_s\} \text{, } E \leftarrow \emptyset \text{, } s_0 \leftarrow x_{s} \text{, } s_{0}.rep \leftarrow x_{s}\text{, } S_{0} \leftarrow {s_1}$\\
    \While{stopping conditions not met}{
        $g_i \leftarrow \textbf{randomGoalIndex}(G)$\\
        $q_{sel} \leftarrow \textbf{bestNearSelection}(V_{a,g_i},\delta_{bn})$\\
        $q_{new} \leftarrow \textbf{propagate}(q_{sel},\tau_{pm})$\\
          \If{\textbf{isPassing}($q_{new}$)}{
            \lIf{$r_{tol}$ radius of $g_{i}$ reached by $q_{sel},q_{new}$}{$g_i \mathrel{{+}{=}} 1$}
            \If{$\textbf{isLocalBest}(q_{new},S_{g_i},\delta_{s})$}{
            $V_{a,g_i} \leftarrow V_{a,g_i} \cup \{q_{new}\}$\\
            $E \leftarrow E \cup \{q_{sel} \rightarrow q_{new,j}\}$\\
            \textbf{pruneNodes}($q_{new},V_{a,g_i},V_{u,g_i}$)\\
            }
          }
    }
\vspace{-0.5em} 
\end{algorithm}

\begin{figure*}[!ht]
   \centering
   \setlength{\tabcolsep}{0.0em}
  \begin{tabular}{ccc}
  \multicolumn{3}{c}{
   \begin{tikzpicture}
      \node[inner sep=0pt,label=right:{\scriptsize sampling-based}] at (0,0) (blue) 
                {\includegraphics[width=.05\textwidth]{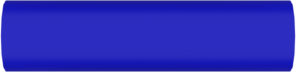}};
      \node[inner sep=0pt,label=right:{\scriptsize polynomial}] at (3.25,0) (yellow) 
                {\includegraphics[width=.05\textwidth]{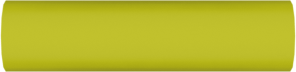}};
      \node[inner sep=0pt,label=right:{\scriptsize search-based}] at (6,0) (green) 
                {\includegraphics[width=.05\textwidth]{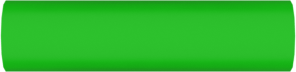}};
      \node[inner sep=0pt,label=right:{\scriptsize start}] at (8.5,0) (black) 
                {\includegraphics[width=.01\textwidth]{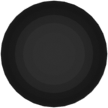}};
      \node[inner sep=0pt,label=right:{\scriptsize goal}] at (10,0) (red) 
                {\includegraphics[width=.01\textwidth]{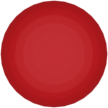}};
   \end{tikzpicture}}   \\[-0.5em]
  \subcaptionbox{Forest\label{fig:maps_forest}}{\includegraphics[height=0.17\linewidth]{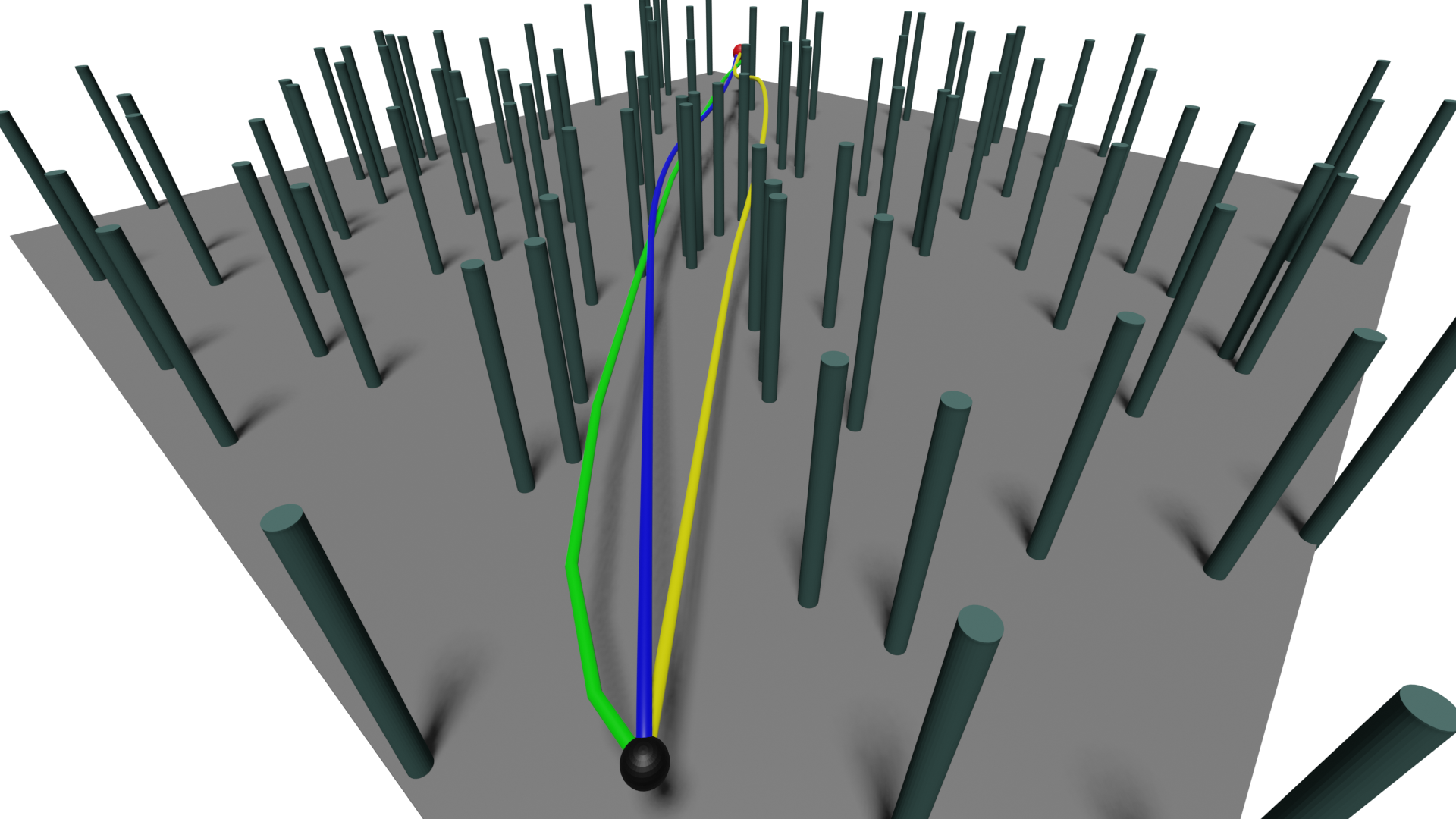}\vspace{-0.5em}}   &
  \subcaptionbox{Office\label{fig:maps_office}}{\includegraphics[height=0.17\linewidth]{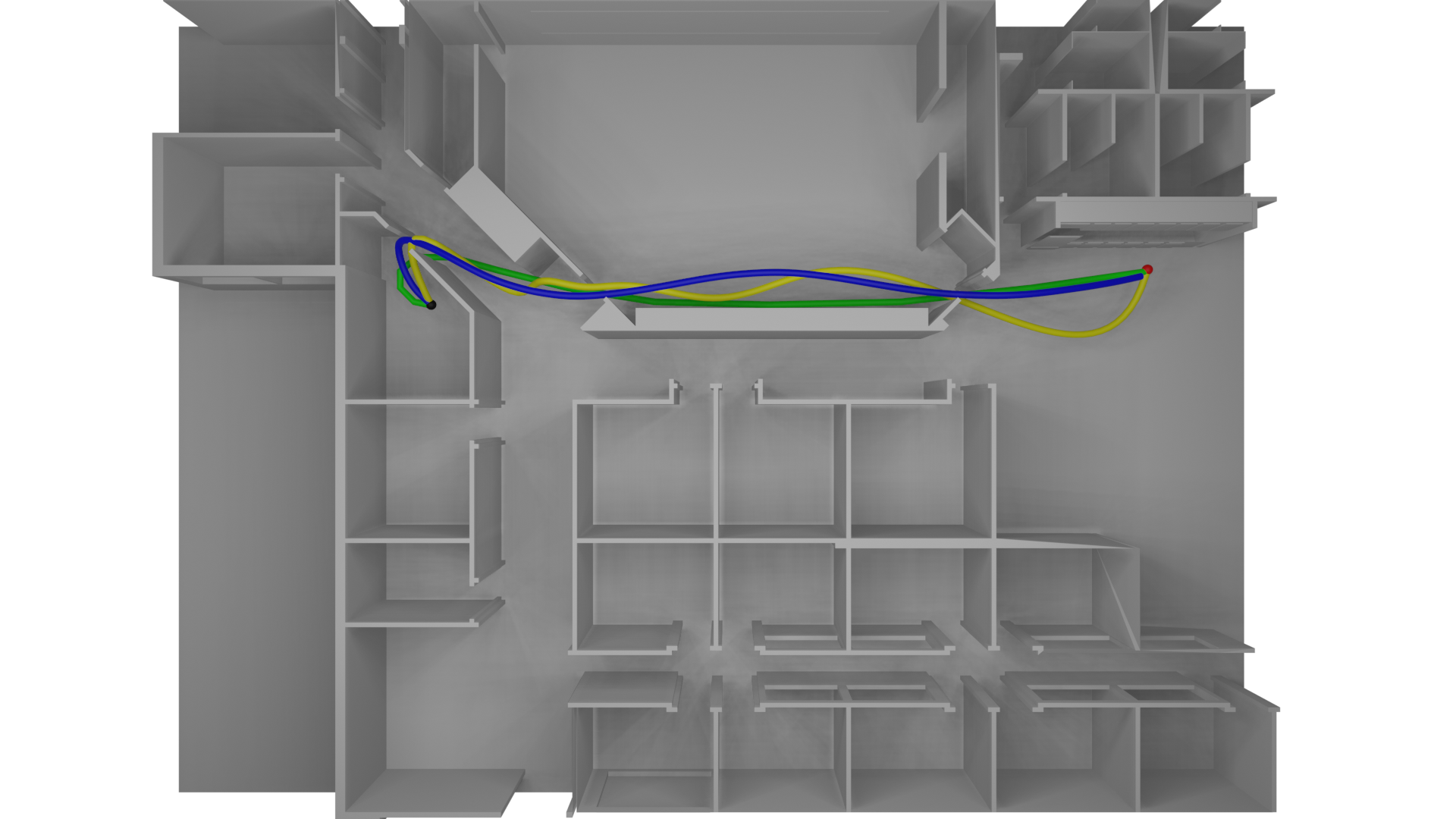}\vspace{-0.5em}}   &
  \subcaptionbox{Racing\label{fig:maps_racing}}{\includegraphics[height=0.17\linewidth]{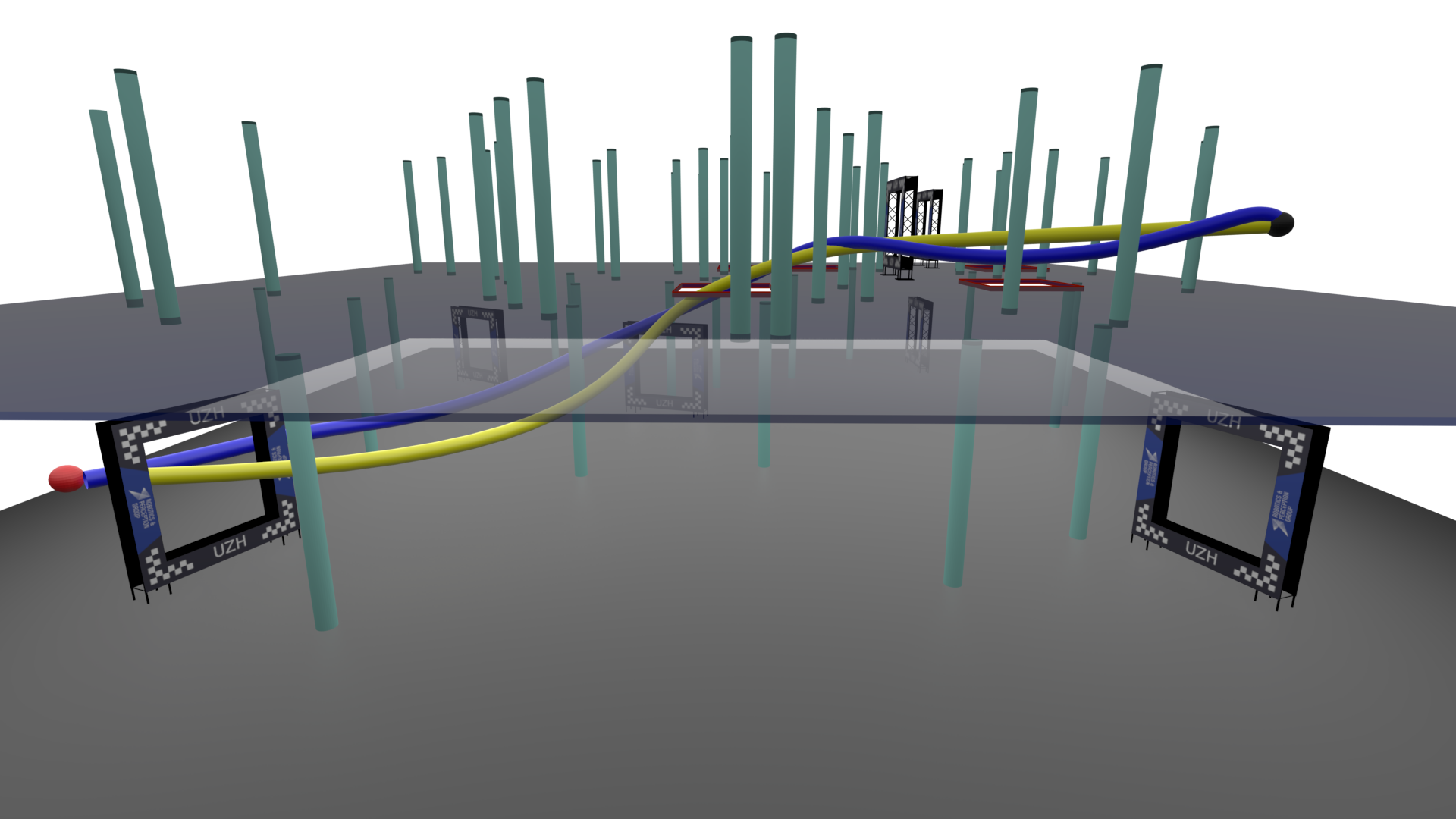}\vspace{-0.5em}}  
   \end{tabular}
   \vspace{-0.2em}
   \caption{Maps of the environments used for evaluating the proposed method and comparison with related algorithms.
   Example trajectories are shown for the proposed sampling-based planner, the polynomial method \cite{richter2016polynomial}, and the search-based method~\cite{Liu_search_based_LQMTC}.
   \label{fig:maps}
   \vspace{-1.5em} 
    }
   \vspace{-0.5em}
\end{figure*}

In each iteration of the Alg.~\ref{alg:multigoal_sst}, a random goal index $g_i$ is selected from the already reached goals.
The \textbf{bestNearSelection} method first generates a state $q_{rand}$ using a random state from the guiding reference between $g_i$ and $g_{i+1}$ with additional uniform random noise with variance $(\sigma^{2}_p,\sigma^{2}_q,\sigma^{2}_v,\sigma^{2}_w)$ for each part of the state.
Selection of a node that lies within $r_{tol}$ distance of goal $g_{i-1}$ is randomly biased with probability $p_g$ to focus exploration around goals.
Then a node $q_{sel}$ with smallest cost from the start within $\delta_{bn}$ radius from $q_{rand}$ is selected for expansion.
If none is found, the nearest node from $V_{a,g_i}$ to $q_{rand}$ is used instead.

The state $q_{sel}$ is further propagated to $q_{new}$ using a 4th-order Runge-Kutta integration of the quadrotor dynamics~\eqref{eq:quat_dyn} for a random time in range $[t_{min},t_{max}]$.
The motor inputs are calculated based on the guiding reference, and the nominal time of applying individual inputs is randomly scaled in the range $[s_{rmin},s_{rmax}]$.
Furthermore, the rotational parts have an additionally randomized rotation axis with variance $\sigma^{2}_{qrot}$.
The randomized time of propagation is a standard way of exploring the solution space by SST, while the randomized scale of reference inputs and the rotation axis is used to account for the artificially added non-continuous rotational parts in the point-mass reference.
The new state $q_{new}$ is checked in the \textbf{isPassing} method to have 1) distance from the reference below a given threshold $\delta_{ref}$, 2) time-cost from the start below a certain ratio $r_{pmm}$ of the closest state in the reference, and 3) collision-free trajectory between $q_{sel}$ and $q_{new}$.
The method thus removes both colliding trajectories and expansions that are far from the point-mass reference in time or in position.
If the goal $g_i$ is reached by the propagated trajectory, the next goal is selected.

Afterwards, method \textbf{isLocalBest} finds nearest witness $s_{new} \in S_{g_i}$ to $q_{new}$. 
If the distance $\norm{q_{new}-s_{new}}$ is above $\delta_{s}$ a new witness $s_{new}$ centered in $q_{new}$ is added to $S_{g_i}$.
The method allows the algorithm to continue if the representative of $s_{new}$ has not been set or the representative's time-cost is lower than the one of the $q_{new}$.
In both cases, $q_{new}$ is added among active nodes $V_{a,g_i}$ and trajectory $q_{sel} \rightarrow q_{new}$ is added among the edges $E$.
Finally, method \textbf{pruneNodes} sets the representative of $s_{new}$ to $q_{new}$ and moves the old representative from $V_{a,g_i}$ to $V_{in,g_i}$.
Furthermore, if the old representative is an inactive leaf node, it is removed from the tree, and the removal is repeated for its parent node recursively until a non-leaf or active node is found.
The algorithm keeps expanding the tree until the stopping conditions are met, consisting of a maximal number of iterations and a limited number of iterations without improvement.

\section{Results\label{sec:results}}

The proposed planning method is evaluated on three environments shown in Figure~\ref{fig:maps}.
It is compared with related algorithms for single and multi-waypoint scenarios.
Furthermore, the method is validated in real-world racing track at over 60km/h in one of the world’s largest motion capture systems ($30\times30\times8m^3$).
The flight is captured in Figure~\ref{fig:illustration}.

All trajectories are found using C++ implementation of the method and a single-core Intel Xeon Gold 6252 CPU with \SI{2.10}{\giga\hertz}.
The parameters of the considered quadrotor and of the algorithms are summarized in Table~\ref{tab:config}.
The stopping conditions used for the method are the maximal number of $2\cdot10^6$ iterations and $2\cdot10^5$ of iterations without improvement.

\vspace{-0.5em}
\begin{table}[!htb] 
   \centering 
   \footnotesize 
   \renewcommand{\tabcolsep}{1.4pt} 
   \renewcommand{\arraystretch}{0.8} 
   \caption{Quadrotor and algorithm parameters\label{tab:config}} 
   \vspace{-1em} 
   \begin{tabular}{c|cc|cc}
     \toprule 
  & Variable & Value & Variable & Value \\[-0.2em] 
       \midrule 
   \parbox[t]{2mm}{\multirow{4}{*}{\rotatebox[origin=c]{90}{Quadrotor}}} & $m$ [\SI{}{\kg}] & 0.85 & $l$ [\SI{}{\meter}] & 0.15 \\
   & $f_{min}$ [\SI{}{\newton}] & 0 & $f_{max}$ [\SI{}{\newton}] & 7 \\
   & $\text{diag}(J)$ [\SI{}{\gram\metre\squared}] & $[1,1,1.7]$ & $\kappa$ [] & 0.05 \\
   & $w_{max}$ [\SI{}{\radian\per\second}] & 15 & & \\[-0.2em] 
   \midrule
   \cite{richter2016polynomial} & $k_{T}$ & $105000$ & $N_{poly}$ & 10 \\[-0.2em] 
   \midrule
   \parbox[t]{2mm}{\multirow{7}{*}{\rotatebox[origin=c]{90}{Method}}} & $d_{tol}$ [\SI{}{\meter}] & 0.3 & $d_c$ [\SI{}{\meter}]& 0.2\\
   & $\delta_{s}$ [] & 0.5 & $\delta_{bn}$ [] & 1.3 \\
   & $(\sigma^{2}_p,\sigma^{2}_q)$[m$^2$,rad$^2$] & $(1.3,0.08)$ & $\sigma^{2}_{qrot}$[rad$^2$] & $0.013$ \\
   & $(\sigma^{2}_v,\sigma^{2}_w)$[$\frac{m^2}{s^2}$,$\frac{rad^2}{s^2}$] & $(8.3,8.3)$ & $[s_{rmin},s_{rmax}]$[] & [0.6,1.4] \\
   & $[t_{min},t_{max}]$[s]  & $[0.004,1.2]$ & $r_{pmm}$[] & 1.05 \\ 
   & $\delta_{ref}$[m] & 2 & $p_g$[] & 0.05 \\[-0.2em] 
\bottomrule 
   \end{tabular}  
   \vspace{-1.0em}
\end{table} 

Computational time of the proposed method is mostly influenced by $\delta_{bn}$ and $\delta_{s}$, that increase the complexity of nearest-neighbor search for larger $\delta_{bn}$ and increase number of created witnesses with smaller $\delta_{s}$.
Values of $(\sigma^{2}_p,\sigma^{2}_q,\sigma^{2}_v,\sigma^{2}_w)$ were scaled to have approximately same scale in encountered states during planning.
The solution quality is mainly influenced by $\delta_{ref}$ and $r_{pmm}$ that bind the quality of the final solution to the quality of the point-mass trajectory.
Other parameters of the method were empirically tuned for the best solution quality.

\subsection{Single goal planning}

\begin{table*}[!htb] 
   \centering 
   \footnotesize 
   {%
   \renewcommand{\tabcolsep}{5.6pt} 
   \renewcommand{\arraystretch}{0.8} 
   \caption{Comparison of the baseline methods and the proposed method in single target scenarios\label{tab:single_target}} 
   \vspace{-1em} 
   \begin{tabular}{lcrrrrrrrcrrrrr}
     \toprule 
    \multirow{2}{*}{Env.} & \multirow{2}{*}{\begin{minipage}{5mm}Test\\case\end{minipage}} & \multicolumn{3}{c}{Polynomial~\cite{richter2016polynomial}} &   \multicolumn{4}{c}{Search-based~\cite{Liu_search_based_LQMTC}} &  \multicolumn{2}{c}{CPC~\cite{foehn2020cpc}} &  \multicolumn{3}{c}{\textbf{Ours}} \\[-0.2em] 
\cmidrule(lr){3-5} \cmidrule(lr){6-9}  \cmidrule(lr){10-11} \cmidrule(lr){12-14}                          &                            & c. time[s] & \multicolumn{1}{c}{$T_{a}$[s]} & $T_{b}$[s] &  c. time[s]$^p$ & $T$[s]$^p$ & c. time[s]$^o$ & $T$[s]$^o$ & c. time[s] & $T$[s] & c. time[s] & \multicolumn{1}{c}{$T_{a}$[s]} & $T_{b}$[s] \\[-0.2em] 
     \midrule 
\multirow{4}{*}{Forest} &0 &0.41 &4.67$\pm$0.63 &3.86 &5.82 &1.80 &10.39 &1.60 &70.23 &\textbf{0.95} &14.92 &1.10$\pm$0.13 &0.96 \\ 
&1 &0.30 &3.43$\pm$0.00 &3.43 &0.73 &1.60 &5.67 &1.40 &67.40 &\textbf{0.96} &2.85 &0.97$\pm$0.00 &\textbf{0.96} \\ 
&2 &0.30 &3.74$\pm$0.93 &3.22 &0.73 &1.40 &1.65 &1.40 &65.49 &\textbf{0.95} &8.03 &0.98$\pm$0.01 &0.96 \\ 
&3 &1.47 &7.20$\pm$1.17 &5.25 &3.85 &2.00 &24.84 &1.80 &- &- &135.83 &1.50$\pm$0.17 &\textbf{1.30} \\[-0.2em] 
\midrule 
\multirow{4}{*}{Office} &0 &1.36 &8.64$\pm$1.03 &7.34 &3.15 &2.80 &22.41 &2.60 &- &- &139.19 &2.38$\pm$0.28 &\textbf{1.93} \\ 
&1 &0.65 &7.50$\pm$0.44 &6.49 &13.18 &2.60 &77.16 &2.20 &- &- &103.64 &1.74$\pm$0.06 &\textbf{1.69} \\ 
&2 &0.89 &9.01$\pm$0.77 &6.38 &11.37 &2.60 &39.70 &2.20 &- &- &155.23 &2.20$\pm$0.13 &\textbf{1.93} \\ 
&3 &0.47 &5.26$\pm$0.35 &5.14 &5.40 &2.40 &35.29 &2.00 &- &- &223.64 &1.81$\pm$0.11 &\textbf{1.58} \\[-0.2em] 
\midrule 
\multirow{4}{*}{Racing} &0 &1.98 &6.56$\pm$0.66 &5.79 &- &- &- &- &- &- &365.91 &1.61$\pm$0.29 &\textbf{1.34} \\ 
&1 &2.06 &6.21$\pm$0.88 &5.13 &- &- &- &- &- &- &428.02 &1.63$\pm$0.15 &\textbf{1.36} \\ 
&2 &1.87 &6.26$\pm$0.42 &4.94 &- &- &- &- &- &- &138.17 &1.45$\pm$0.12 &\textbf{1.37} \\ 
&3 &1.91 &5.27$\pm$0.69 &4.73 &- &- &- &- &- &- &604.55 &2.14$\pm$0.74 &\textbf{1.57} \\[-0.2em] 
\bottomrule 
   \end{tabular} 
   } 
   \vspace{-2.0em}
\end{table*}

The first set of experiments is for single target scenarios where a trajectory is planned between start and goal states only.
This allows comparison between all considered baseline algorithms.
The first baseline algorithm is the polynomial method~\cite{richter2016polynomial} which jointly minimizes snap and final time of reaching the goal (using time penalty $k_T$) with polynomials of order $N_{poly}$.
Contrary to the original method, we employ the PRM* algorithm to find paths between individual waypoints.
The second baseline is the search-based minimum-time method~\cite{Liu_search_based_LQMTC}, where we limit the point-mass state to include position derivatives only up to acceleration due to computational complexity.
Method~\cite{Liu_search_based_LQMTC} is evaluated for pessimistic \SI{22.2}{\meter\per\second\squared} and for optimistic (though unfeasible) \SI{31.4}{\meter\per\second\squared} maximal acceleration per axis, denoted with `p' and `o', respectively.
The first is is derived from the maximal motor thrust in diagonal lateral motion, while the second as maximal acceleration in lateral motion along single axis.
Finally, the time-optimal CPC optimization method~\cite{foehn2020cpc} is extended to a variant for cluttered environments using position constrained by the ESDF map of the environment.

The results for the single goal planning are summarized in Table~\ref{tab:single_target} for all three environments with four test cases (i.e., different start-goal positions) for each environment.
We report the computational time of each algorithm and the time duration (i.e., solution quality) $T$ of found trajectories.
For the polynomial method and the proposed method, we show the average duration with standard deviation $T_{a}$ and the best duration $T_{b}$ found within 30 conducted runs per test case.
The results show that the proposed method is able to find the highest quality trajectories for the majority of the test cases.
The computational times of our methods, however, render it only as an offline planning method similarly to \cite{foehn2020cpc,Liu_search_based_LQMTC}.
The polynomial method, although fast in finding solutions, is by far the one with the highest trajectory durations.
This demonstrates the inherent inability of the polynomial trajectory planning to represent the minimum-time trajectories.
The search-based method is 14\%-66\% worse in solution quality for even the optimistic (`o') acceleration limit than the proposed sampling-based method and has comparable computational times.
The comparably lower solution quality of the search-based method is caused jointly by the time discretization and the per-axis acceleration limit that restricts exploitation of the full quadrotor actuation.
The search-based method is able to find solutions only for the Forest and Office scenarios with 2D planning problems.
In the Racing environment, where planning in 3D is required, the method is not able to find a solution before reaching the memory limit of \SI{32}{\giga\byte}.
This is caused by the curse of dimensionality of the state discretization which requires exponential growth of memory for additional searched dimensions.
On the other hand, the time-optimal CPC method is able to find a solution only for the three easiest scenarios in the Forest environment.
In other environments, the CPC is not able to find a feasible solution due to the introduced collision constraints that add non-convexities that the optimization-based method can not handle.
However, the comparison with the solutions found by CPC shows the very small gap in duration between time-optimal solutions and the best trajectories found by the proposed method.
Furthermore, our method is faster approximately by a factor of ten compared to the CPC.

\subsection{Multi-waypoint planning}

The second set of evaluations focuses on the multi-waypoint scenarios.
The scalability of the method is tested for a variable number of targets and density of the obstacles in the environment.
Furthermore, the related algorithms are compared for the multi-waypoint drone-racing task in both environments with and without obstacles.

Computation time and solution quality are shown in Table~\ref{tab:multi_target} to demonstrate the scalability of the method in multi-waypoint planning.
It is tested in an environment similar to the Forest~\ref{fig:maps_forest} with an increasingly larger number of randomly placed columns and targets (NT).
The number of tested columns $[50,100,150,200]$ corresponds, respectively, to [3.5\%, 6.6\%, 10\%, 13\%] portion of occupied environment.

\begin{table}[!htb] 
   \centering 
   \footnotesize 
   {\renewcommand{\tabcolsep}{4.6pt} 
   \renewcommand{\arraystretch}{0.8} 
   \caption{Scalability evaluation with increasing number of waypoints and obstacles\label{tab:multi_target}} 
   \vspace{-1em} 
   \begin{tabular}{lrrrrrrrr}
     \toprule 
 \multirow{2}{*}{NT} &  \multicolumn{2}{c}{N. Col. 50} & \multicolumn{2}{c}{N. Col. 100} & \multicolumn{2}{c}{N. Col. 150} & \multicolumn{2}{c}{N. Col. 200}  \\[-0.2em]  
\cmidrule(lr){2-3}\cmidrule(lr){4-5}\cmidrule(lr){6-7}\cmidrule(lr){8-9} 
  & c.t.[s] & T[s]  & c.t.[s] & T[s]  & c.t.[s] & T[s]  & c.t.[s] & T[s]  \\[-0.2em] 
       \midrule 
2 &26.5&0.74 &88.8&0.74 &158.2&0.79 &220.6&0.76\\ 
3 &155.9&1.90 &389.2&1.98 &524.9&2.54 &1044.2&2.79\\ 
4 &248.0&2.90 &421.1&3.40 &491.3&4.38 &-&-\\ 
5 &1095.8&4.77 &958.1&4.88 &1125.5&6.38 &-&-\\[-0.2em] 
\bottomrule 
   \end{tabular} 
   } 
   \vspace{-0.5em}
\end{table}

The presented data shows the expected trend of higher computational times and larger trajectory durations for both higher number of targets and obstacles.
For the highest number of columns and targets, the method was unable to find a solution within the limited number of iterations.
This is mainly due to a high number of narrow passages that have a low probability of being sampled, especially for high-speed multi-waypoint planning.

The multi-waypoint planning is also tested in a racing environment (Fig.~\ref{fig:maps_racing}) that features seven gates.
Our method is compared with the polynomial trajectory planning and the CPC.
The search-based method is not considered as it allows only single target planning.
Both trajectory durations and computational times are presented in Table~\ref{tab:multi_target_arena} for environments with and without obstacles.

\begin{table}[!htb] 
\vspace{-0.5em}
   \centering 
   \footnotesize 
   {\renewcommand{\tabcolsep}{2.5pt} 
   \renewcommand{\arraystretch}{0.8} 
   \caption{Comparison for the drone racing task\label{tab:multi_target_arena}} 
   \vspace{-1em} 
   \begin{tabular}{lrrrrrrrr}
     \toprule 
    \multirow{2}{*}{O.}  &  \multicolumn{3}{c}{Poly.~\cite{richter2016polynomial}} &  \multicolumn{2}{c}{CPC~\cite{foehn2020cpc}} &  \multicolumn{3}{c}{\textbf{Ours}} \\[-0.2em]  
\cmidrule(lr){2-4} \cmidrule(lr){5-6}  \cmidrule(lr){7-9}      &  c.t.[s] & \multicolumn{1}{c}{$T_{a}$[s]} & $T_{b}$[s] & c.t.[s] & $T$[s] & c.t.[s] & \multicolumn{1}{c}{$T_{a}$[s]} & $T_{b}$[s] \\[-0.2em]  
       \midrule 
w/o &4.34 &15.33$\pm$0.00 &15.33 &1009.34 &\textbf{6.53} &919.84 &6.99$\pm$0.06 &6.88 \\ 
w &8.12 &27.54$\pm$0.62 &26.98 &- &- &734.65 &7.10$\pm$0.06 &\textbf{7.01} \\[-0.2em] 
\bottomrule 
   \end{tabular} 
   } 
   \vspace{-0.5em}
\end{table}

Similar to the single target scenarios, the solution quality of the polynomial trajectories is significantly lower.
Comparison of the CPC and our method shows similar computational time.
Our method is able to find solutions with only $5.3$\% higher trajectory duration than the optimal CPC.
Most importantly, unlike the CPC, our method can find the trajectory for the cluttered environment while the solution quality is comparable to the one without obstacles.

The planned trajectory has been finally validated in real-world flights in multi-waypoint racing track depicted in Figure~\ref{fig:maps_racing}.
The used quadrotor platform is based on the open-hardware and open-software Agilicious quadrotor framework~\cite{agilicious} with the same parameters as listed in Table~\ref{tab:config}.
A Model Predictive Control with Incremental Nonlinear Dynamic Inversion low-level controller~\cite{sun2021comparative} were used to follow the planned trajectory.
Figure~\ref{fig:illustration} and the attached video show the flight of successfully executed trajectory reaching velocities over 60km/h.
The root-mean-square error of tracking the trajectory is 0.55m, which is similar to the ones measured in~\cite{foehn2020cpc,agilicious} for the same platform. 
This shows that our method is able to plan feasible high-speed and near time-optimal trajectories in cluttered environments.

\section{Conclusions\label{sec:conclusion}}

This paper introduced a novel hierarchical, sampling-based method that plans with an incrementally more complex quadrotor model. 
The method is used to find minimum-time trajectories over a given sequence of waypoints in cluttered environments, which is a problem that had not been previously solved in its entirety.
We showed that the proposed method outperforms all existing baselines in finding minimum-time trajectories for cluttered environments and achieves near time-optimal solutions for scenarios without obstacles.
The planned trajectories were validated in a real-world racing track at speeds over 60km/h.
For future work, we plan to include perception-awareness to allow onboard gate detection and visual-inertial odometry in drone racing.
We want to investigate optimality guarantees of our method and further study methods suitable for minimum-time online planning.
Finally, we believe the method could be extended for different vehicles as a general framework for hierarchical planning with increasing model fidelity.

\bibliographystyle{IEEEtran}
\bibliography{main}

\end{document}